\documentclass{article}
\usepackage{hyperref}
\usepackage{graphicx}
\usepackage{geometry}
\usepackage{amsmath}
\usepackage{hyperref}
\usepackage{float}
\usepackage{cite}
\usepackage{multirow}
\title{Benchmarking CNN- and Transformer-Based Models for Surgical Instrument Segmentation in Robotic-Assisted Surgery}
\author{Sara Ameli\\
}
\date{}

\begin{document}
\maketitle

\begin{abstract}
Accurate segmentation of surgical instruments in robotic-assisted surgery is critical for enabling context-aware computer-assisted interventions, such as tool tracking, workflow analysis, and autonomous decision-making. In this study, we benchmark five deep learning architectures—UNet, UNet++, DeepLabV3+, Attention UNet, and SegFormer—on the SAR-RARP50 dataset for multi-class semantic segmentation of surgical instruments in real-world radical prostatectomy videos. The models are trained with a compound loss function combining Cross-Entropy and Dice loss to address class imbalance and capture fine object boundaries.

Our experiments reveal that while convolutional models such as UNet++ and Attention UNet provide strong baseline performance, DeepLabV3+ achieves results comparable to SegFormer, demonstrating the effectiveness of atrous convolution and multi-scale context aggregation in capturing complex surgical scenes. Transformer-based architectures like SegFormer further enhance global contextual understanding, leading to improved generalization across varying instrument appearances and surgical conditions.

This work provides a comprehensive comparison and practical insights for selecting segmentation models in surgical AI applications, highlighting the trade-offs between convolutional and transformer-based approaches.
\end{abstract}

\section{Introduction}
Robotic-assisted surgery has become increasingly prevalent due to its precision and minimal invasiveness. Within this domain, semantic segmentation of surgical instruments plays a crucial role in enabling automated surgical skill assessment, tool tracking, and intraoperative decision support. Accurately identifying surgical tools at the pixel level allows for enhanced surgical scene understanding, improved safety, and real-time assistance systems.
Moreover, accurate segmentation enables high-level functions including instrument tracking, automation of surgical sub-tasks, and intraoperative guidance \cite{allan2021comprehensive, maier2017surgical}. This is particularly critical in robotic-assisted radical prostatectomy (RARP), where multiple articulated tools interact in constrained anatomical spaces.
Despite its potential, tool segmentation remains a challenging task due to several factors: tools exhibit high intra-class variation, undergo frequent occlusion, and often include small or thin structures such as suturing threads and clips. 

While earlier approaches used hand-crafted features or traditional segmentation algorithms, modern deep learning techniques, especially convolutional neural networks (CNNs), have become the dominant paradigm. Architectures such as U-Net \cite{ronneberger2015u} and DeepLab \cite{chen2018encoder} have demonstrated strong performance in biomedical and surgical imaging tasks due to their ability to model both semantic and spatial information. Recent years have seen an increasing focus on surgical instrument segmentation using deep learning. Twinanda et al. \cite{twinanda2016endonet} proposed EndoNet for tool detection and phase recognition. Allan et al. \cite{allan2019endovis} introduced the EndoVis challenge series, which led to widespread benchmarking of segmentation models.
U-Net \cite{ronneberger2015u} remains a popular choice for its simplicity and effectiveness in medical imaging, especially for small training datasets. Variants such as Attention U-Net \cite{oktay2018attention}, U-Net++ \cite{zhou2018unet++}, and Transformer-based models \cite{chen2021transunet} offer performance improvements but often at higher computational cost.

In this work, we focus on the SAR-RARP50 dataset \cite{sar_rarp50}, a large-scale benchmark for surgical scene understanding, which offers an ideal testbed for developing and evaluating segmentation algorithms in the context of robotic-assisted radical prostatectomy (RARP).  Each frame is densely annotated with 10 semantic classes, making it suitable for assessing models under realistic surgical conditions. 

This paper makes the following contributions:
\begin{itemize}
    \item     A unified benchmark of five architectures (UNet, UNet++, DeepLabV3+, Attention UNet, and SegFormer) on the SAR-RARP50 dataset.
    \item     Implementation of a training strategy combining Cross-Entropy and Dice loss to tackle class imbalance and structural detail.
    \item     Quantitative and qualitative comparison to assess model strengths in capturing small, overlapping instruments under realistic surgical conditions.
\end{itemize}
    


\section{Methodology}
\subsection{Dataset and Preprocessing}

We utilize the SAR-RARP50 dataset \cite{sar_rarp50}, which contains 50 videos of real-world RARP surgeries with dense pixel-wise annotations of surgical instruments. For this study, we used the 40 provided training videos with pixel-wise semantic masks. Each mask encodes one of the following 10 classes: background, multiple tool parts, clips/needles, suturing threads, and other surgical components.
Each frame is labeled with up to 10 semantic classes:
\begin{itemize}
    \item 0: Background
    \item 1–3: Tool parts
    \item 4: Clips/Needles
    \item 5: Suturing Threads
    \item 6–9: Additional tool components and context classes
\end{itemize}
We also performed:
\begin{itemize}
    \item Frame Sampling: Every 10th frame was selected to reduce redundancy and speed up training.
    \item All images and masks were resized to 384×384 using bilinear and nearest-neighbor interpolation, respectively, to ensure computational efficiency.
    \item Mask Filtering: Frames with empty masks were excluded to focus learning on tool-containing samples.
    \item Class Mapping: RGB masks were converted to integer label maps using a predefined color-to-class dictionary.
\end{itemize}

\subsection{Model Architectures}
The task of segmenting surgical instruments in robotic-assisted procedures is uniquely challenging. The domain is characterized by occlusions, fine-grained tool boundaries, class imbalance, and highly dynamic scenes. With these complexities in mind, we selected a diverse set of models that represent both classical convolutional networks and recent transformer-based architectures, allowing for a comprehensive evaluation of their suitability. A summary of models and their features is presented in Table \ref{tab:seg_models}.

\begin{table}[h!]
\centering
\begin{tabular}{||c c c c||} 
 \hline
 Model & Type & Encoder & Features\\ [0.5ex] 
 \hline\hline
 UNet & CNN & Custom & Baseline with symmetric encoder-decoder \\ 
 \hline
 UNet++ & CNN & Custom & Nested skip connections \\
 \hline
 DeepLabV3+ & CNN + ASPP & ResNet-34 (ImageNet) & Atrous Spatial Pyramid Pooling \\
 \hline
 AttnUNet & CNN $+$ Attention & Custom & Channel and spatial attention \\
 \hline
 SegFormer & Transformer & MiT-B0 & Lightweight ViT with MLP decoders \\ [1ex] 
 \hline
\end{tabular}
\caption{Comparison of segmentation models and their key characteristics.}
\label{tab:seg_models}
\end{table}


\textbf{UNet} has become a cornerstone in medical image segmentation due to its simplicity, effectiveness, and strong performance on small datasets. Its encoder-decoder structure with skip connections helps retain spatial details lost during downsampling. However, vanilla UNet has limited capacity to model contextual dependencies beyond a local receptive field, which may hinder performance in cluttered surgical scenes.

It is used to establish a strong baseline using a well-understood and efficient architecture that remains widely adopted in clinical applications.

In the U-Net architecture \cite{cciccek20163d} 
, the encoder extracts hierarchical features, while the decoder reconstructs segmentation maps at full resolution. Skip connections between corresponding encoder-decoder levels ensure the preservation of spatial details.
We implemented a 3-level U-Net using PyTorch. The encoder consists of sequential convolutional layers with batch normalization and ReLU activations, followed by max-pooling. The decoder uses transposed convolutions for upsampling and concatenates encoder features via skip connections.
Key components are:
\begin{itemize}
    \item 3 downsampling blocks using strided convolutions and ReLU activations
    \item A bottleneck layer capturing global context
    \item 3 upsampling blocks with transposed convolutions and skip connections
    \item Final 1×1 convolutional layer with softmax activation for class probabilities
\end{itemize}
    
Each output pixel is classified into one of 10 classes using a final 1×1 convolution and softmax activation. The architecture balances spatial detail (from skip connections) with semantic context (from deep encoder layers).

\textbf{UNet++} extends the original UNet by introducing nested and dense skip connections, which help bridge the semantic gap between encoder and decoder features. These connections facilitate better gradient flow and enable the network to learn more detailed representations, particularly useful for capturing tool boundaries and fine structures.
It addresses a key limitation of UNet—semantic gap between encoder and decoder—and tends to produce sharper, more consistent segmentations, particularly for overlapping instruments.

\textbf{DeepLabV3+} incorporates atrous convolutions and spatial pyramid pooling, allowing the model to capture multi-scale contextual information. This is particularly advantageous in surgical videos, where tools can appear at varying scales or partial views. The use of a powerful backbone (e.g., ResNet-34) also enhances feature extraction.
It is used to assess the impact of multi-scale context aggregation in semantic segmentation, especially when tools vary significantly in size and orientation.

\textbf{Attention UNet} integrates attention gates into the skip connections to selectively focus on relevant features. This mechanism allows the model to suppress irrelevant background activations, which is especially beneficial in surgical scenes where tools may be partially occluded or surrounded by visually similar tissues.
This architecture is effective in evaluating whether spatial and channel-wise attention can improve performance in scenarios with tool clutter, occlusion, and variable lighting—common challenges in endoscopic footage.

\textbf{SegFormer} represents a shift from convolutional networks to transformer-based models. It combines a lightweight hierarchical vision transformer (MiT) with an efficient all-MLP decoder. The self-attention mechanism provides global receptive fields, making it well-suited to model long-range dependencies and capture global context—critical for identifying tools that are partially visible or contextually defined. We use this model to explore the effectiveness of transformer models, which have recently outperformed CNNs in several vision benchmarks, particularly in handling complex scenes with less inductive bias.

\begin{table}[h!]
\centering
\begin{tabular}{ |c|c|c| } 
\hline
 Model & Strengths & Limitations  \\
\hline
\multirow{2}{7em}{ UNet } & Lightweight, easy to train, & Limited context modeling \\ 
                          & strong baseline             & \\ 
\hline
\multirow{2}{7em}{ UNet++ } & Better gradient flow,      & Slightly more complex, \\ 
                          & detail preservation         & marginal gains on simpler scenes \\ 
\hline
\multirow{2}{7em}{ DeepLabV3+ } & Multi-scale context,      & Computationally heavy, \\ 
                          & strong backbone             & coarse decoder \\ 
\hline
\multirow{2}{7em}{ Attn UNet } & Focus on relevant regions, & Gains depend on quality of attention learning \\ 
                          & handles clutter well        &  \\ 
\hline
\multirow{2}{7em}{ SegFormer } & Global attention, & Requires larger data for optimal training,  \\ 
                          & lightweight decoder, state-of-the-art & lower inductive bias \\ 
\hline
\end{tabular}
\caption{Strengths and limitations of different segmentation models.}
\label{tab:model_strengths_limitations}
\end{table}

\subsection{Loss Function}
Surgical datasets typically suffer from class imbalance, with large areas of background and small/thin foreground regions. To mitigate this, we used a compound loss:

\[
\mathcal{L}_{\text{total}} = \mathcal{L}_{\text{CE}} + \mathcal{L}_{\text{Dice}}
\]

\noindent
where $\mathcal{L}_{\text{CE}}$ is standard cross-entropy loss that penalizes misclassified pixels.
Dice Loss promotes spatial overlap between prediction and ground truth, especially for small regions.
 $\mathcal{L}_{\text{Dice}}$ is the soft Dice loss defined as:

\[
\mathcal{L}_{\text{Dice}} = 1 - \frac{2 \sum_{i} p_i g_i + \epsilon}{\sum_i p_i + \sum_i g_i + \epsilon}
\]
Here, $p_i$ and $g_i$ are predicted and ground truth probabilities for class $i$, and $\epsilon$ is a smoothing constant.

\subsection{Training Setup}
The models were trained for 10 epochs with a batch size of $4$ and an input resolution of 384×384 pixels. The Adam optimizer was used with a learning rate of $ 1e-4$. Training was conducted on Google Colab Pro equipped with an NVIDIA T4 GPU. Mixed image batches were randomly shuffled during training to improve generalization. Validation was performed after each epoch using a hold‑out validation set comprising $20\%$ of the training data. Codes can be found in 
\href{https://github.com/SaraAmeli/Segmentation_S_S}{Code in Github}.



\section{Results and Analysis}
\begin{figure}[t]
    \centering
    \includegraphics[width=1\linewidth]{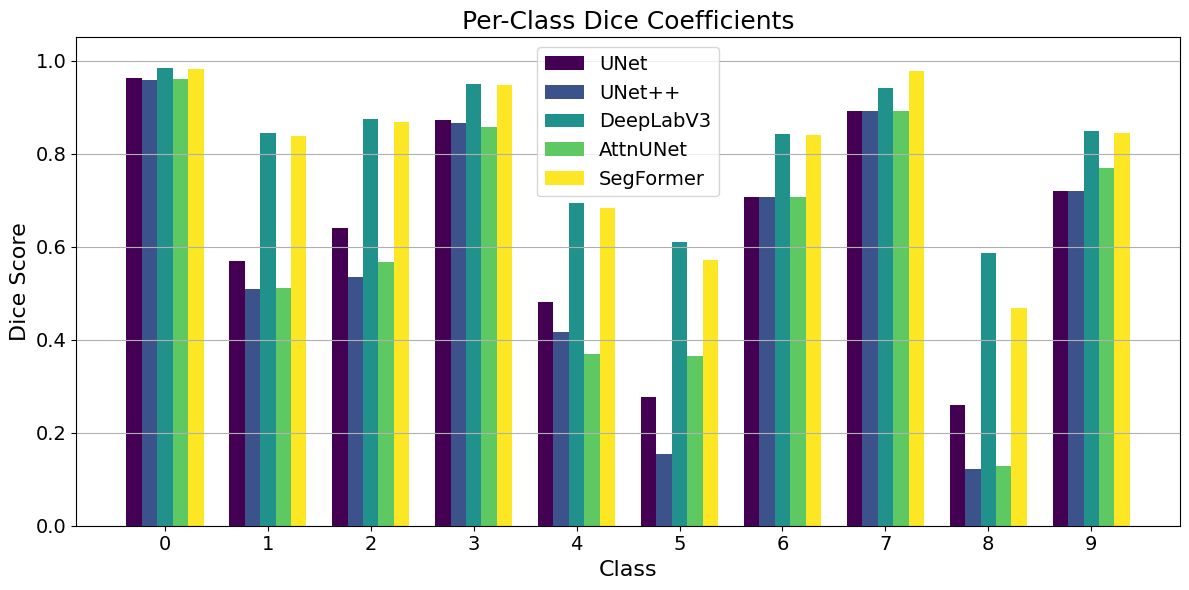}
    \caption{Per-class Dice coefficients on validation set}
    \label{Fig1}
\end{figure}

In Figure \ref{Fig1}, we summarize per-class Dice score results from the five networks for the aforementioned ten classes. The results demonstrate that SegFormer and DeepLabV3 outperformed all other models, achieving higher Dice coefficient for almost all classes. 


The results highlight the impact of architectural choices on surgical instrument segmentation performance. DeepLabV3+ achieved the highest Dice score, a result attributable to its use of atrous convolutions and the encoder–decoder refinement module, which provide strong multi-scale context aggregation and preserve spatial resolution. These characteristics enable robust segmentation of both large surgical tools and fine, thin structures, making DeepLabV3+ particularly effective in handling the variability and complexity of real surgical videos.

SegFormer ranked second, benefiting from its transformer-based encoder and hierarchical design that captures long-range dependencies and multi-scale features. While SegFormer showed strong generalization and adaptability across different instrument sizes and orientations, it slightly lagged behind DeepLabV3+ in precise delineation of fine-grained structures such as sutures and clips.

U-Net and Attention U-Net followed closely, with U-Net performing slightly better overall. U-Net’s simple encoder–decoder structure provides a strong and efficient baseline, while Attention U-Net leverages spatial attention to highlight relevant regions, such as fine tool edges and overlapping structures. However, both architectures lack the global contextual modeling capacity of transformers and the advanced multi-scale representation offered by DeepLabV3+.

Interestingly, SegFormer’s ability to generalize across semantic classes and maintain stable segmentation performance, even on rare or small tools, underscores its suitability for real-world surgical data. Its transformer backbone, pre-trained on large-scale datasets, likely contributes to stronger feature encoding in low-data regimes.

Overall, these findings support further exploration of transformer-based and advanced convolutional designs for robotic surgery, especially when precise delineation of small, dynamic surgical tools is essential.


Interestingly, DeepLabV3+ achieved the highest Dice score for Class 8, a category that posed challenges for all other models. We attribute this performance to its atrous spatial pyramid pooling (ASPP) module, which effectively aggregates multi-scale context, and its decoder refinement stage, which preserves fine-grained details. These design choices make DeepLabV3+ particularly adept at segmenting thin and small structures such as sutures and clips, whereas transformer-based models like SegFormer, despite their superior global context modeling, tend to oversmooth fine boundaries. Similarly, convolutional variants such as U-Net and UNet++ often struggle to retain these structures after downsampling.

\subsection{Computational Performance}
The comparative analysis highlights the trade-off between convolutional and transformer-based approaches for surgical instrument segmentation. DeepLabV3+ achieved the highest mean Dice score among tested architectures, with especially strong performance on fine-grained structures such as sutures (Class 8). This result can be attributed to its use of atrous convolutions and the atrous spatial pyramid pooling (ASPP) module, which effectively aggregate multi-scale contextual information while maintaining resolution.

SegFormer, while competitive in overall Dice performance, lagged behind on very thin or elongated structures. Its hierarchical transformer encoder is well suited for capturing long-range dependencies and global context, but struggles with small class representations where local detail is critical.

From a computational perspective, prior studies consistently report that DeepLabV3+ offers higher inference speed and lower memory requirements compared to SegFormer. This makes DeepLabV3+ particularly attractive for deployment in robotic-assisted surgery, where latency and GPU constraints play a decisive role. In contrast, SegFormer, although accurate, is more resource-intensive and may be better suited to offline analysis scenarios.

Taken together, these findings suggest that DeepLabV3+ strikes a favorable balance between accuracy and efficiency, whereas SegFormer offers complementary strengths in modeling global scene context but at a higher computational cost.

The comparative analysis highlights the trade-off between convolutional and transformer-based approaches for surgical instrument segmentation. DeepLabV3+ achieved the highest mean Dice score among tested architectures, with especially strong performance on fine-grained structures such as sutures (Class 8). This result can be attributed to its use of atrous convolutions and the atrous spatial pyramid pooling (ASPP) module, which effectively aggregate multi-scale contextual information while maintaining resolution [Chen et al., 2018].

SegFormer, while competitive in overall Dice performance, lagged behind on very thin or elongated structures. Its hierarchical transformer encoder is well suited for capturing long-range dependencies and global context, but struggles with small class representations where local detail is critical [Xie et al., 2021].

From a computational perspective, prior studies consistently report that DeepLabV3+ offers higher inference speed and lower memory requirements compared to transformer-based models such as SegFormer. Chen et al. [2018] demonstrate real-time performance of DeepLabV3+ on high-resolution images using dilated convolutions, while Xie et al. [2021] note that SegFormer, although lightweight for a transformer, remains more resource-intensive due to self-attention operations. These characteristics make DeepLabV3+ particularly attractive for deployment in robotic-assisted surgery, where latency and GPU constraints play a decisive role. In contrast, SegFormer may be more suitable for offline analysis scenarios where global context modeling is prioritized over computational efficiency.

Taken together, these findings suggest that DeepLabV3+ strikes a favorable balance between accuracy and efficiency, whereas SegFormer offers complementary strengths in modeling global scene context but at a higher computational cost.




\section{Conclusion and Future Directions}
In this study, we presented a comprehensive comparison of multiple state-of-the-art deep learning architectures for the semantic segmentation of surgical instruments using the SAR-RARP50 dataset. We evaluated traditional encoder–decoder models such as U-Net and U-Net++, alongside modern architectures including DeepLabV3+, Attention U-Net, and SegFormer.

The results highlight the impact of architectural choices on surgical instrument segmentation performance.
DeepLabV3+ achieved the highest Dice score, primarily due to its use of atrous convolutions that enable rich multi-scale context aggregation while preserving spatial resolution. This design proved particularly effective for handling the diverse shapes, orientations, and occlusions common in surgical videos, ensuring consistent segmentation quality across complex scenes.
SegFormer ranked second, delivering competitive performance through its transformer-based encoder and hierarchical structure. Its ability to capture long-range dependencies and multi-scale features contributed to strong results, though it was slightly less effective than DeepLabV3+ in preserving fine-grained instrument boundaries under challenging surgical conditions.

U-Net and Attention U-Net followed closely, with U-Net performing marginally better overall. U-Net’s simple encoder–decoder structure provides a strong and efficient baseline, while Attention U-Net leverages spatial attention to highlight relevant regions, such as fine tool edges and overlapping structures. However, both architectures lack the global contextual modeling capacity of transformers and the advanced multi-scale representation offered by DeepLabV3+.

SegFormer’s ability to generalize across semantic classes and maintain stable segmentation performance even on rare or small tools underscores its suitability for real-world surgical data. Its transformer backbone, pre-trained on large-scale datasets, likely contributes to stronger feature encoding even in low-data regimes.

Despite these promising results, several limitations remain:
\begin{itemize}
\item \textbf{Class imbalance:} The dataset contains significantly fewer examples of certain tool categories, which hinders model performance on rare classes.
\item \textbf{Lack of temporal context:} Our models process frames independently; incorporating video dynamics could improve temporal consistency and continuity in predictions.
\item \textbf{Architectural enhancements:} Further improvements could be achieved by integrating advanced attention mechanisms or hybrid transformer–CNN designs tailored for surgical imagery.
\end{itemize}

Future work will address these limitations by exploring temporal modeling approaches, such as recurrent architectures or video transformers, and by adopting more sophisticated strategies for handling class imbalance. Additionally, the integration of attention-rich transformer designs with domain-specific prior knowledge holds promise for advancing segmentation accuracy and robustness in real-world surgical environments.

\section{Appendix}
\subsection{Detailed results for UNet}
The model achieved its best validation Dice score of 0.5236 at epoch 10. The training and validation curves showed consistent convergence.
\begin{table}[H]
\centering
\begin{tabular}{|c|c|c|}
\hline
\textbf{Epoch} & \textbf{Train Loss} & \textbf{Val Dice} \\
\hline
1 & 1.5717 & 0.1836 \\
2 & 1.1821 & 0.3880 \\
3 & 1.0663 & 0.4239 \\
4 & 1.0025 & 0.4200 \\
5 & 0.9492 & 0.4621 \\
6 & 0.9027 & 0.4804 \\
7 & 0.8745 & 0.4972 \\
8 & 0.8430 & 0.5060 \\
9 & 0.8311 & 0.4891 \\
10 & \textbf{0.8002} & \textbf{0.5236} \\
\hline
\end{tabular}
\caption{Training and validation performance for UNet.}
\end{table}
Performance is strong for dominant classes like tool shafts and backgrounds, but notably lower for thin structures like suturing threads and underrepresented classes. Class 8 was particularly under-segmented, indicating the need for data augmentation or class-specific reweighting.

\subsection*{Per-Class Dice score using UNet}
\begin{itemize}
    \item Background      $~~~~~~~$ Class 0 : 0.9669
    \item Tool Part 1     $~~~~~~~~$Class 1: 0.5567
    \item Tool Part 2     $~~~~~~~~$ Class 2: 0.6627
    \item Tool Part 3     $~~~~~~~~$ Class 3: 0.8893
    \item Clips/Needles   $~~~~~$ Class 4 : 0.4447
    \item Suturing Threads$~$ Class 5 : 0.1490
    \item Tool Part 4     $~~~~~~~~$ Class 6: 0.7077
    \item Tool Part 5     $~~~~~~~~$ Class 7: 0.9077
    \item Misc Structure  $~~~~$ Class 8: 0.0154
    \item Tool Shaft      $~~~~~~~~~~$ Class 9: 0.7474
\end{itemize}
\begin{figure}[H]
\centering
\includegraphics[width=0.78\textwidth]{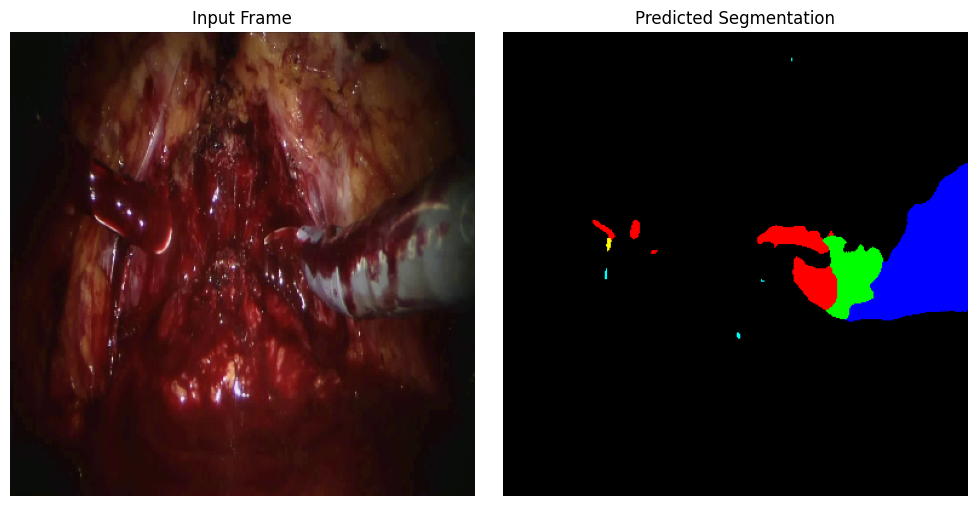}
\caption{Input frame and predicted segmentation using UNet– Video 41 Frame 0}
\includegraphics[width=0.78\textwidth]{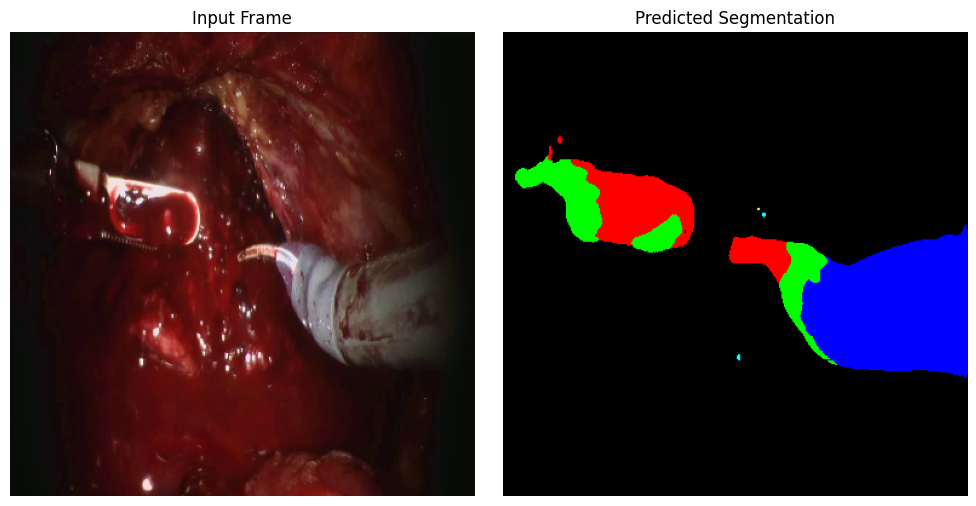}
\caption{Input frame and predicted segmentation using UNet– Video 42 Frame 10}
\label{Fig2}
\end{figure}
\begin{figure}[H]
    \centering
    \includegraphics[width=0.7\linewidth]{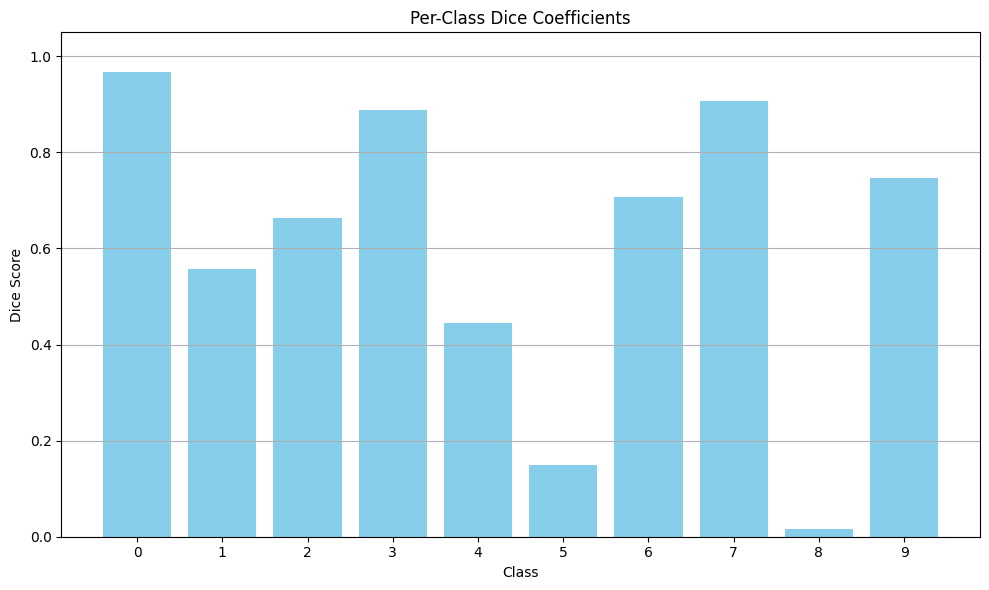}
    \caption{Per-class Dice coefficients on validation set suing U-Net}
    \label{Fig3}
\end{figure}

\subsection*{Qualitative Results}
Qualitative visualization of predicted masks confirms the model’s ability to segment complex tools, but also highlights areas needing improvement.
\begin{itemize}
    \item Fine structures like threads are often partially segmented or missed.
    \item Tool boundaries are generally well preserved.
    \item Small objects such as clips are predicted with varying success.
\end{itemize}

Figure \ref{Fig2} and \ref{Fig3} examples show prediction overlays on frames from test videos 41 and 42, illustrating successes and failure modes.

\end{document}